\documentclass[a4paper, 11pt]{article}

\usepackage{fullpage}
\usepackage{graphicx}
\usepackage{wrapfig}
\usepackage{natbib}
\usepackage[english]{babel}
\usepackage[center]{caption}
\usepackage{mathtools}
\usepackage{mathptmx}   
\usepackage{latexsym}          
\usepackage{url}
\usepackage{multirow}
\usepackage{amssymb}
\usepackage{amsthm}
\usepackage{amssymb}
\usepackage{amsmath}
\usepackage{units}
\usepackage{mathrsfs}
\usepackage{setspace}

\begin{document}

\title{An Experiment on Using Bayesian Networks for Process Mining}
\author{Catarina Moreira}

\date{ }

\maketitle

\doublespace

\begin{abstract}

Process mining is a technique that performs an automatic analysis of business processes from a log of events with the promise of understanding how processes are executed in an organisation.

Several models have been proposed to address this problem, however, here we propose a different approach to deal with uncertainty. By uncertainty, we mean estimating the probability of some sequence of tasks occurring in a business process, given that only a subset of tasks may be observable.

In this sense, this work proposes a new approach to perform process mining using Bayesian Networks. These structures can take into account the probability of a task being present or absent in the business process. Moreover, Bayesian Networks are able to automatically learn these probabilities through mechanisms such as the maximum likelihood estimate and EM clustering.

Experiments made over a Loan Application Case study suggest that Bayesian Networks are adequate structures for process mining and enable a deep analysis of the business process model that can be used to answer queries about that process.

\end{abstract}

\section{Introduction}

Process mining is a technique that enables the automatic analysis of business processes based on event logs. Instead of designing a workflow, process mining consists in gathering the information of the tasks that take place during the workflow process and storing that data in structured formats called the event logs~\citep{Aalst11}. While gathering this information, it is assumed that (1) each event refers to a task in the business process, (2) each event is associated to an instance of the workflow and (3) since the events are stored by their execution time, it is assumed that they are sorted~\citep{Aalst04}.

During the last decade, process mining has been growing a lot of attention in the scientific community due to its promise to provide techniques for process discovery that will lead to an increase of productivity and to the reduction of costs~\citep{Aalst05}.

Process modelling can be seen as the techniques to graphically represent a business process. This graphical representation describes dependencies between activities that need to be executed together in order to fulfil a business target~\citep{Weske12}.

Since in process mining the order of the events is taken into consideration, there are already many models that can be directly applied to represent the workflow. Some of those models include Markov Chains~\citep{Ferreira07,Rebugea12}, Petri Nets~\citep{Aalst98}, Neural Networks~\citep{Cook98} and BPMN~\citep{Aalst11}. However, Markov Chains and Petri Nets are the models that are most used in the literature of process mining~\citep{Tiwari08}.

In this work, it is proposed an alternative representation of business process by using Bayesian Networks. A Bayesian Networks can be defined as an acyclic directed graph in which each node represents a random variable and each edge represents a direct influence from the source node to the target node (conditional dependencies)~\citep{Spirtes01}. They differ from Markov Chains, because of their cycle-free and directed structure. Moreover, Bayesian Networks have the advantage of dealing with uncertainty differently from Markov Chains. In the latter, business processes are modelled as a chain of events that are observed to occur. Under a Bayesian Network perspective, this does not apply: each task can either be \emph{present} or \emph{absent} in the business process. Therefore, it is possible to perform special analysis that will enable the computation of the probability of some task of the business process occurring, given that we do not know which tasks have already been performed~\citep{Pearl09}.

With this research work, we argue that the capabilities of Bayesian Networks provide a promising technique to model business processes, to perform analysis regarding risk management, cost reduction, finding irrelevant / repetitive tasks, etc.

The outline of this work is as follows. Section~\ref{sec:mc} presents a brief summary of Markov Chains. Section~\ref{sec:bn} makes an introduction to Bayesian Networks. It shows how to compute probabilistic inferences and presents some learning techniques that are used to automatically learn conditional probabilities in Bayesian Networks. Section~\ref{sec:bn_process} presents how Bayesian Networks can be applied in the realm of process mining. This section demonstrates how one can define the structure of a Bayesian Network and how one can perform automatic learning. Section~\ref{sec:case_study} presents a case study in which we apply the proposed network. Finally, Section~\ref{sec:conclusions}  summarises the current work, presents the main conclusions achieved and some directions for future work.


\section{Markov Chains}\label{sec:mc}

A Markov Chain is defined by a state space $Val(X)$ and a model that defines, for every state $x \in Val( X )$ a next-state distribution over $Val(X)$. More precisely, the transition model $\tau$ specifies for each pair of states $x,x'$ the probability $\tau( x \rightarrow x')$ of going from state $x$ to $x'$~\citep{Koller09}.

\begin{figure}[h!]
\centering
\includegraphics[scale = 0.5]{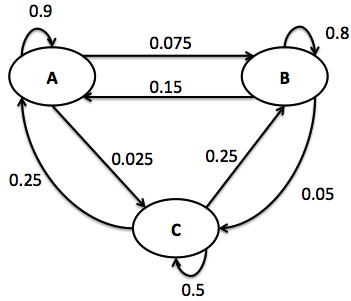}
\caption{Example of a Markov Chain}
\label{fig:markov_ex}
\end{figure}

In Markov Chains, the transition probability matrix must be stochastic, that is, each row of the matrix must sum to one. Matrix~\ref{eq:transition} represents the transition matrix of the Markov Chain in Figure~\ref{fig:markov_ex}.
\begin{equation}
P_{transition} = \left[ \begin{matrix} ~0.9	& 0.075	&0.025~\\
							   0.15	& 0.8	& 0.05	\\
							  0.25	& 0.25	& 0.5	\\
			 \end{matrix} \right]  
			\label{eq:transition}
\end{equation} 

Suppose that one is in state $B$ at time $n$. In order to compute the evolution of the system for $ n + 1$, one just needs to perform a matrix multiplication between the current state and the transition probability matrix. The current state $B$ will be encoded as vector $\left[ 0~1~0  \right]$.

\begin{equation}
 \left[ \begin{matrix} 0 & 1 & 0 \end{matrix} \right]   \left[ \begin{matrix} ~0.9	& 0.075	&0.025~\\
							   								0.15	& 0.8	& 0.05	\\
							 							 0.25	& 0.25	& 0.5	\\
											 \end{matrix} \right] = \left[ \begin{matrix} 0.15 & 0.8 & 0.05 \end{matrix} \right] 
\label{eq:transition_ex}
\end{equation}

The calculations in formula~\ref{eq:transition_ex} show that the probability from transiting from state $B \rightarrow A$ is $0.15$. The probability of transiting from $B \rightarrow B$ is $0.8$. And the probability of transiting from state $B \rightarrow C$ is $0.05$.

Moreover, if one wishes to compute the probability of the sequence $A \rightarrow B \rightarrow B \rightarrow C$, one would need to perform the following calculations:
\begin{equation}
Pr( A \rightarrow B \rightarrow B \rightarrow C ) = Pr( A \rightarrow B ) Pr( B \rightarrow B  ) Pr( B \rightarrow C ) = 0.075 \times 0.8 \times 0.05 = 0.003 
\end{equation}

\section{Bayesian Networks}\label{sec:bn}

Bayesian Networks are directed acyclic graphs in which each node represents a different random variable from a specific domain and each edge represents a direct influence from the source node to the target node~\citep{Pearl97}. The graph represents independence relationships between variables and each node is associated with a conditional probability table (CPT) which specifies a distribution over the values of a node given each possible joint assignment of values of its parents. The full joint distribution of a Bayesian Network, where $X$ is the list of variables, is given by~\citep{russel10}:

\begin{equation}
Pr_c( X_1, \dots, X_n ) = \prod_{i=1}^n  Pr( X_i | Parents(X_i) )
\label{eq:joint}
\end{equation}

The formula for computing classical exact inferences on Bayesian Networks is based on the full joint distribution (Equation~\ref{eq:joint}). Let $e$ be the list of observed variables and let $Y$ be the remaining unobserved variables in the network. For some query $X$, the inference is given by:

\begin{equation}
Pr_c( X | e ) = \alpha Pr_c(X, e) = \alpha \left[ \sum_{y \in Y} Pr_c( X, e, y) \right]
\label{eq:inference}
\end{equation}
\[ \text{Where~~~}\alpha = \frac{1}{ \sum_{x \in X} Pr_c(X = x, e) } \]
The summation is over all possible $y$, i.e., all possible combinations of values of the unobserved variables $y$. The $\alpha$ parameter, corresponds to the normalisation factor for the distribution $Pr(X | e)$~\citep{russel10}.

\subsection{Example of Application}

\begin{figure}[h!]
\centering
\includegraphics[scale=0.6]{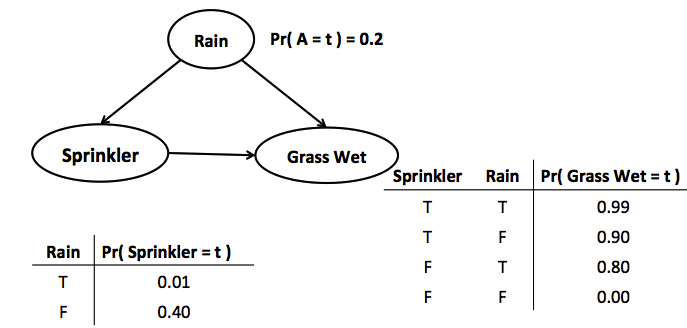}
\caption{Example of a Bayesian Network.}
\label{fig:example_bn}
\end{figure}

Consider the Bayesian Network in Figure~\ref{fig:example_bn}. Suppose that we want to determine the probability of raining given that we know that the grass is wet.
In order to perform such inference on a Bayesian Network, one can use Equation~\ref{eq:inference} in the following way:
\begin{equation}
Pr(~R~=~T~| W = T~)~=\alpha~Pr(~R~=~T~)\times \sum_{ s \in S} Pr(~S = s~|~R =~T~) \times Pr(~W~=~T~|~S~=~s,~R~=~T~)
\label{eq:pr_r}
\end{equation}
\begin{equation}
\begin{split}
Pr(~R~=~T~| W = T~) = \alpha~0.2 \times [ Pr( S = T | R = T ) Pr( W = T | S = T, R = T ) + \\ + Pr( S = F | R = T) Pr( W = T | S = F, R = T ) ]
\end{split}
\end{equation}
\begin{equation}
Pr(~R~=~T~| W = T~) = \alpha~0.2 \times \left[ 0.01 \times 0.99 + 0.99 \times 0.8  \right] = \alpha~0.1604 = 0.3577 
\label{eq:pr_r_final}
\end{equation}
Given that Bayesian Networks are based on the Na\"{i}ve Bayes rule, one needs to normalize the final probabilities by a factor $\alpha$. This normalisation factor $\alpha$ corresponds to:
\begin{equation}
\alpha = \frac{1}{Pr( R = T | W = T ) + Pr( R = F | W = T ) }
\label{eq:norm_fact}
\end{equation}
So, in order to compute $\alpha$, one also needs to compute the probability of not raining given that the grass is wet, Pr( R = F $|$ W = T ):
\begin{equation}
Pr(~R~=~F~| W = T~)~=\alpha~Pr(~R~=~F~)\times \sum_{ s \in S} Pr(~S = s~|~R =~F~) \times Pr(~W~=~T~|~S~=~s,~R~=~F~)
\label{eq:pr_nr}
\end{equation}
\begin{equation}
\begin{split}
Pr(~R~=~F~| W = T~) = \alpha~0.8 \times [ Pr( S = T | R = F ) Pr( W = T | S = T, R = F ) +\\ + Pr( S = F | R = F) Pr( W = T | S = F,~R~=~F~ ) ]
\end{split}
\end{equation}
\begin{equation}
Pr(~R~=~F~| W = T~) = \alpha~0.8 \times \left[ 0.4 \times 0.9 + 0.6 \times 0  \right] = \alpha~0.288 
\label{eq:pr_nr_final}
\end{equation}
Going back to the normalisation factor in Equation~\ref{eq:norm_fact}, one can substitute $Pr(R=T | W = T)$ by the result in Equation~\ref{eq:pr_r_final} and $Pr( R = F | W = T )$ by the results in Equation~\ref{eq:pr_nr_final}.
\begin{equation}
\alpha = \frac{1}{ 0.1604 + 0.288 } = \frac{1}{0.4484}
\end{equation}
Now that we have computed the normalisation factor, the final probabilities are:
\begin{equation}
Pr(~R~=~T~| W = T~) = \alpha~0.1604 = 0.3577 
\end{equation}
\begin{equation}
Pr(~R~=~F~| W = T~) = \alpha~0.288  = 0.6423
\end{equation}

\subsection{Learning in Bayesian Networks}

There are two main approaches to build a Bayesian Network. One is to construct the network by hand and to use the knowledge of an expert to estimate the conditional probability tables. The second is to use statistical models to automatically \emph{learn} these probabilities~\citep{Koller09}.

Estimating the conditional probabilities by hand with the knowledge of an expert is problematic for several reasons. In some situations, the network is so big that it is almost impossible for the expert to make a reliable assignment of the probabilities to the random variables. Moreover, in many situations, the distribution of the data varies according to its application and through time. This makes it impossible for an expert to reliably estimate the probabilities associated to the random variables of the Bayesian Network.

Statistical models, on the other hand, offer a mechanism to automatically learn a model that represents the probability distribution of some population.

According to the situation that one is modelling, one can have a fully observed dataset or have an incomplete dataset (or partially observed). For the scope of this work, we will only address the problem of learning in Bayesian Networks with a fully observed dataset and a known graphical structure. The data are considered fully observed if on  each of the training instances there is a full instantiation to all the random variables of our sample space~\citep{Murphy12}.

\subsubsection{Maximum Likelihood Estimation in Bayesian Networks} \label{sec:mle}

The maximum likelihood estimation is a statistical method that assumes that data follows a Gaussian probability distribution. The mean and the variance of the probability distribution can be estimated by only knowing a partial sample of the dataset~\citep{Bishop07}.

\begin{figure}[h!]
\centering
\includegraphics[scale=0.6]{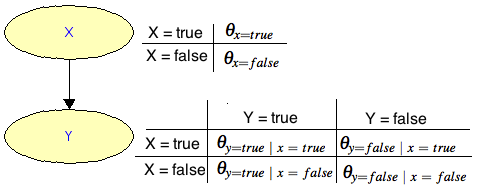}
\caption{Example of a Bayesian Network structure with unspecified conditional probability tables.}
\label{fig:learning_ex1}
\end{figure}

Suppose that we have a Bayesian Network just like specified in Figure~\ref{fig:learning_ex1}. This network is parameterized by a parameter vector $\theta$ which specifies the parameters for the conditional probability distribution of the network.

The training instances regarding Figure~\ref{fig:learning_ex1} consist in a tuple of the form $\langle x\left[ m \right], y\left[ m \right]  \rangle$, where $x$ is an instance of the random variable $X$, $y$ is an instance of the random variable $Y$ and $m$ is the $m_{th}$ training example from the training dataset $D$ of size $M$.

The likelihood function is given by:
\begin{equation}
L( \theta : D ) = \prod_{m=1}^M Pr( x\left[ m \right], y\left[ m \right] : \theta )
\label{eq:learn1}
\end{equation}

Since in a Bayesian Network we can specify a full joint probability distribution $Pr( x\left[ m \right], y\left[ m \right] : \theta )$ by the chain rule, then, Equation~\ref{eq:learn1} becomes:

\begin{equation}
L( \theta : D ) =   \prod_{m} Pr( x\left[ m \right] : \theta_X ) Pr( y\left[ m \right] | x\left[ m \right]  : \theta_{Y|X} )
\label{eq:learn2}
\end{equation}
\begin{equation}
L( \theta : D ) =   \left( \prod_{m} Pr( x\left[ m \right] : \theta_X ) \right) \left( \prod_{m} Pr( y\left[ m \right] | x\left[ m \right]  : \theta_{Y|X} ) \right)
\label{eq:learn3}
\end{equation}

Equation~\ref{eq:learn3} shows that the likelihood function can be decomposed into two separate terms. If we had $N$ random variables, then Equation~\ref{eq:learn3} would also have $N$ terms. Each of these terms is called a \emph{local likelihood function} and can estimate how well a variable can predict its parents.

Moreover, one can expand the second term of Equation~\ref{eq:learn3} for each instance of $x$ in the following way:
\begin{equation}
\begin{split}
\prod_{m} Pr( y\left[ m \right] | x\left[ m \right]  : \theta_{Y|X} ) = \prod_{m : x\left[ m \right] = x_{true}} Pr( y\left[ m \right] | x\left[ m \right]  : \theta_{Y|x_{true}} )\prod_{m : x\left[ m \right] = x_{false}} Pr( y\left[ m \right] | x\left[ m \right]  : \theta_{Y|x_{false}} )
\end{split}
\label{eq:learn4}
\end{equation}


Going back to the simple Bayesian Network in Figure~\ref{fig:learning_ex1}, if we analyse the first term of~Equation\ref{eq:learn4}, we can see that it refers to the number of instances of the training data in which $x=true$. This gives us two sets: $x=true , y=true$ and $x=true, y=false$. Equation~\ref{eq:learn6} discriminates these instances.
\begin{equation}
 \prod_{m : x\left[ m \right] = x_{true}} Pr( y\left[ m \right] | x\left[ m \right]  : \theta_{Y|x_{true}} ) = \theta_{y=true | x = true} . \theta_{y = false | x = true}
\label{eq:learn6}
\end{equation}

Then, Equation~\ref{eq:learn6} becomes:
\begin{equation}
\theta_{y = false | x = true} = \frac{count( \langle x=true, y=false \rangle )}{ count( \langle x=true, y=false \rangle ) + count( \langle x=true, y=true \rangle ) }
\end{equation}
\begin{equation}
\theta_{y = false | x = true} = \frac{count( \langle x=true, y=false \rangle )}{ count( \langle x=true \rangle ) }
\label{eq:learn7}
\end{equation}

From Equation~\ref{eq:learn7}, we can see that the maximum likelihood estimate for a Bayesian Network with a known structure and fully observed data consists in simply counting how many times each of the possible assignments of $X$ and $Y$ appear in the training data. In order to obtain a probability value, we normalize this score by counting the total number of instances that class $X$ appears.

\subsection{SamIam}

SamIam - \emph{Sensitivity, Analysis, Modeling, Inference and More} - is a tool that enables the graphical modeling of Bayesian Networks. It was developed by the \emph{Automated Reasoning Group} form the University of California\footnote{\url{http://reasoning.cs.ucla.edu/samiam/}}.

SamIam is composed of a graphical interface and a reasoning engine. The graphical interface provides an easy way to model Bayesian Networks by specifying the random variables as nodes, causal connections as edges and the respective conditional probability tables. The reasoning engine, on the other hand, can perform classical inferences over the plotted Bayesian Network, make parameter estimations by learning mechanisms, sensitivity analysis, etc. For the scope of this work, only the classical inference and the learning mechanisms will be necessary.

\begin{figure}[h!]
	\parbox{.3\linewidth}{
	\centering
	\includegraphics[scale=0.4]{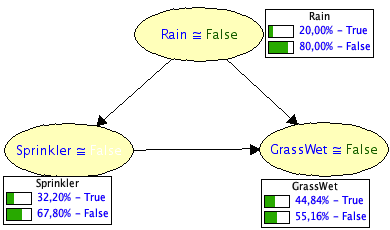}
	\caption{SamIam representation of the Bayesian Network of Figure~\ref{fig:example_bn}.}
	\label{fig:ex1}
	}
	\hfill
	\parbox{.3\linewidth}{
	\centering
	\includegraphics[scale=0.4]{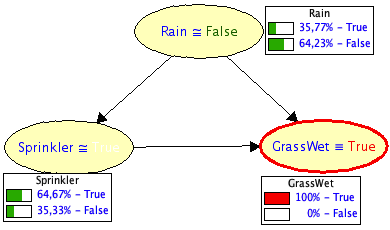}
	\caption{Example of SamIam's inference engine: $Pr(R|W = T)$, $Pr( S | W = T )$.}
	\label{fig:ex2}	
	}
	\hfill
	\parbox{.3\linewidth}{
	\centering
	\includegraphics[scale=0.4]{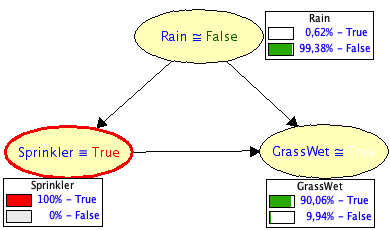}
	\caption{Example of SamIam's inference engine: $Pr(R|S = T)$, $Pr( W | S = T )$.}
	\label{fig:ex3}	
	}
\end{figure}

\begin{figure}[h!]
	\parbox{.3\linewidth}{
	\centering
	\includegraphics[scale=0.4]{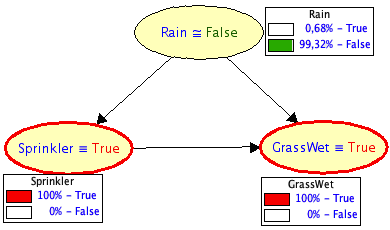}
	\caption{Example of SamIam's inference engine: $Pr(~R|W = T, S=T~)$.}
	\label{fig:ex4}
	}
	\hfill
	\parbox{.3\linewidth}{
	\centering
	\includegraphics[scale=0.4]{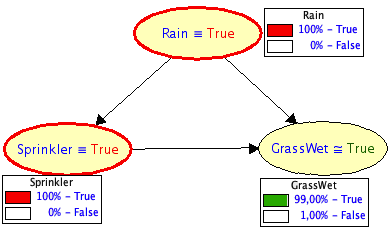}
	\caption{Example of SamIam's inference engine: $Pr(~W |S = T, R = T~)$.}
	\label{fig:ex5}	
	}
	\hfill
	\parbox{.3\linewidth}{
	\centering
	\includegraphics[scale=0.4]{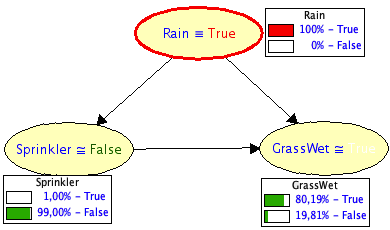}
	\caption{Example of SamIam's inference engine: $Pr(~S|R = T~)$, $Pr(~W~|~R = T~)$.}
	\label{fig:ex6}	
	}
\end{figure}

Examples of SamIam's graphical interface are given by Figures~\ref{fig:ex1} to~\ref{fig:ex6}.

In Figure~\ref{fig:ex1}, it is presented the Bayesian Network from Figure~\ref{fig:example_bn} under the SamIam graphical interface. The marginal probabilities for each node are automatically computed as soon as the user builds the Bayesian Network. Figure~\ref{fig:ex1} shows that: $Pr( R = T ) = 0.2$, $Pr( R = F) = 0.8$, $Pr( S = T ) = 0.3220$, $Pr( S = F ) = 0.6780$, $Pr( W = T ) = 0.4484$, $Pr( W = F ) = 0.5516$.

Figure~\ref{fig:ex2} represents a graphical representation  of the inference that was manually computed in Equations~\ref{eq:pr_r_final} and~\ref{eq:pr_nr_final}. The red markers represent variables which are \emph{observed}. That is, variables, which have occurred. They can be seen as the conditions of probabilities. For instance, in the manually computed probability in Equation~\ref{eq:pr_r}, the observed variable was the condition $W = T$, that is, we are asking the probability of Raining given that it was observed that the grass was wet.

For large Bayesian Networks, the inference process becomes very heavy and hard to be computed manually. Therefore, SamIam provides an easy interface that automatically performs such heavy operations.

In process mining, event logs are usually associated with a large amount of tasks, which can be mapped into nodes of a Bayesian Network. Consequently, for the scope of this work, we chose the capabilities of SamIam to automatically compute inferences related to the probability of certain sequences of tasks occurring. This mechanism will be more detailed in Section~\ref{sec:samiam} of this work.

\section{Bayesian Networks for Process mining}\label{sec:bn_process}

Probabilistic graphical models, such as Bayesian Networks, are usually used for probabilistic inferences, that is, asking queries to the model and receiving answers in the form of probability values.

Under the realm of process mining, Bayesian Networks can represent activities as nodes (i.e. random variables) and the edges between activities can be seen as transitions between these tasks. From this structure, it is possible to automatically learn the conditional probability tables from a complete log of events using the Maximum Likelihood Estimations (Section~\ref{sec:mle}). If the log is incomplete, then a Bayesian Network can also automatically learn and estimate the probability tables through the usage of EM Clustering, just like used in the work of~\cite{Bobek13}, who developed a Bayesian Network to recommend business processes.

In the literature, business processes that are learnt from event logs are usually represented by either Markov Chains or Petri Nets~\citep{Weske12}. In this work, however, we propose another approach to model business processes using Bayesian Networks. The reason why we do this is concerned with the fact that Bayesian Networks can deal with uncertainty more easily.

Bayesian Networks provide advantages in situations where we do not know if some task has occurred and we need to determine the probability of the process terminating or the probability of the process reaching some other task. Therefore, these structures provide more insights when there are high levels of uncertainty when compared to Markov Chains. 	

\subsection{Defining the Strucuture}

\begin{figure}[h!]
	\parbox{.5\linewidth}{
	\centering
	\includegraphics[scale=0.3]{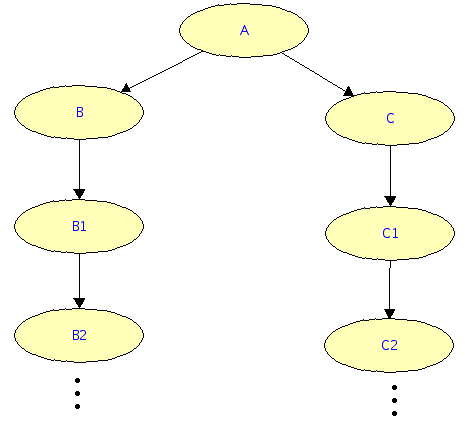}
	\caption{Example of a representation of a Bayesian Network with cycles.}
	\label{fig:cycle_example}
	}
	\hfill
	\parbox{.5\linewidth}{
	\centering
	\includegraphics[scale=0.3]{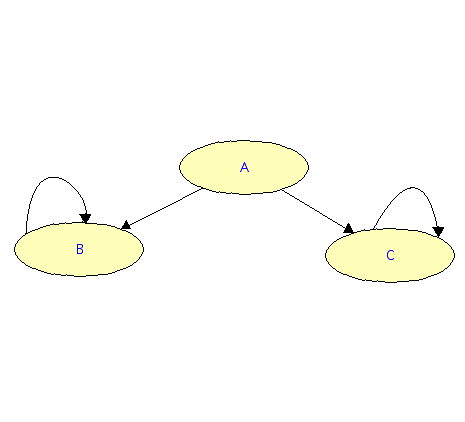}
	\caption{Example of a representation of a Markov Chain with cycles.}
	\label{fig:cycle_example_markov}	
	}
\end{figure}

Another advantage of Bayesian Networks is that they allow the direct representation of business process diagrams by capturing the direct dependencies between tasks.
However, they do not allow an explicit representation of cycles, because Bayesian Networks are directed acyclic graphs. To represent a cycle, in a Bayesian Network, one would need to create many instances of the same node, which is intractable to perform inferences, since the inference problem is \emph{NP-Complete} (Figures~\ref{fig:cycle_example} and~\ref{fig:cycle_example_markov}).

\begin{figure}[h!]
	\parbox{.5\linewidth}{
	\centering
	\includegraphics[scale=0.4]{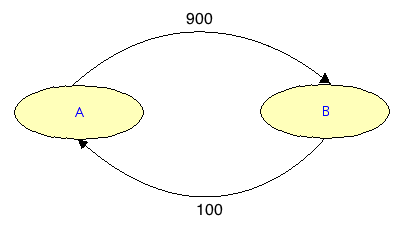}
	\caption{Example of a Markov Chain with a cycle.}
	\label{fig:cycles_h1}
	}
	\hfill
	\parbox{.5\linewidth}{
	\centering
	\includegraphics[scale=0.4]{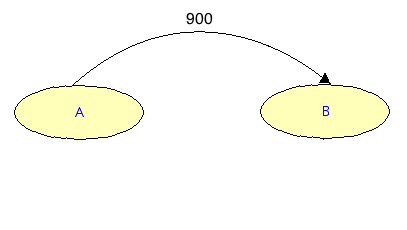}
	\caption{Conversion of the Markov Chain to a Bayesian Network by removing the weakest edge.}
	\label{fig:cycles_h2}	
	}
\end{figure}

In this work, in order to eliminate cycles from the log of events, we used an heuristic that would choose the most probable transitions between nodes. For instance, suppose that there is a transition from nodes $A \rightarrow B$ that occurred 900 times. Suppose also that there is a transition from nodes $B \rightarrow A$ that occurred 100 times. Following the proposed heuristic, we would only represent the Bayesian Network with the transition $A \rightarrow B$. Figures~\ref{fig:cycles_h1} and~\ref{fig:cycles_h2} illustrates this example.

\begin{figure}[h!]
\centering
\includegraphics[scale=0.5]{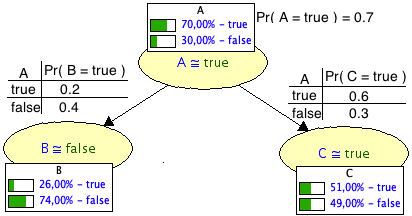}
\caption{Example of a Bayesian Network with no mutually exclusive nodes.}
\label{fig:mutual1}
\end{figure}

Another structure that Bayesian Networks cannot represent directly is concerned with mutual exclusion. Two events are mutually exclusive if they cannot occur at the same time. Bayesian Networks can capture mutually exclusive events through the notions of independence by manually adding new edges to the network. For instance, consider the business process represented by the  Bayesian Network in Figure~\ref{fig:mutual1}. Nodes $B$ and $C$ represent the end of the process, while node $A$ represents a task that begins the process. In this situation, and following the semantics of the business process, it is required that nodes $B$ and $C$ become mutually exclusive. That is, the process flow can only end in one of these nodes and not on both of them at the same time.

\begin{figure}[h!]
	\parbox{.45\linewidth}{
	\centering
	\includegraphics[scale=0.5]{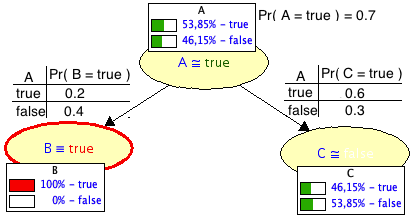}
	\caption{Example of a Bayesian Network with no mutually exclusive nodes.}
	\label{fig:mutual2}	
	}
	\hfill
	\parbox{.45\linewidth}{
	\centering
	\includegraphics[scale=0.5]{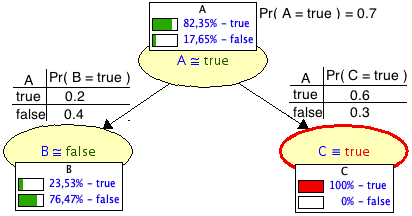}
	\caption{Example of a Bayesian Network with no mutually exclusive nodes.}
	\label{fig:mutual3}	
	}
\end{figure}

\begin{figure}[h!]
\centering
\includegraphics[scale=0.35]{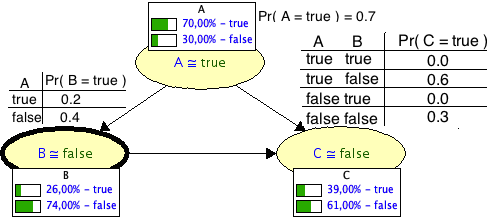}
\caption{Example of a Bayesian Network with mutual exclusion.}
\label{fig:mutual4}
\end{figure}

\begin{figure}[h!]
	\parbox{.45\linewidth}{
	\centering
	\includegraphics[scale=0.5]{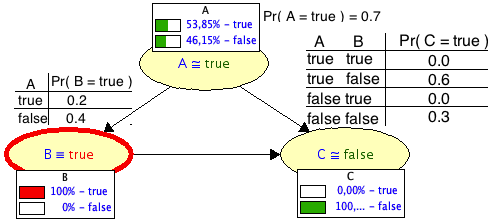}
	\caption{Example of a Bayesian Network with mutual exclusion.}
	\label{fig:mutual5}	
	}
	\hfill
	\parbox{.45\linewidth}{
	\centering
	\includegraphics[scale=0.5]{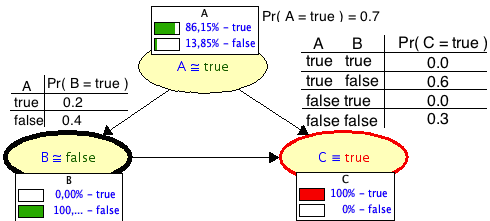}
	\caption{Example of a Bayesian Network with mutual exclusion.}
	\label{fig:mutual6}	
	}
\end{figure}

As one can see in Figures~\ref{fig:mutual2} and~\ref{fig:mutual3}, the Bayesian Network cannot represent this mutual exclusion. When computing Bayesian Inferences, all nodes depend on each other. Therefore, in order to semantically represent \emph{node B cannot occur at the same time as node C}, one needs to add an extra edge between $B \rightarrow C$. This additional edge will create a new dependency between these nodes. One can manually configure the conditional probability table of node $C$ to represent this mutually exclusion: when node $B$ is set to $true$, then the probability of occurring $C$ is zero and vice-versa. The mutual exclusion of the Bayesian Network in Figure~\ref{fig:mutual1} is illustrated in Figures~\ref{fig:mutual4} to~\ref{fig:mutual6}.

Note that, in Figures~\ref{fig:mutual4} to~\ref{fig:mutual6}, the probability of node $C$ occurring when nothing is observed changed when compared to the Bayesian Network of Figure~\ref{fig:mutual1}. This happened, because of the extra edge that was added in the later Bayesian Network, which ended up changing the configurations of the conditional probability tables and, consequently, final probability values.

\subsection{SamIam: Designing a Bayesian Network}\label{sec:samiam}

\begin{figure}[h!]
\includegraphics[scale=0.5]{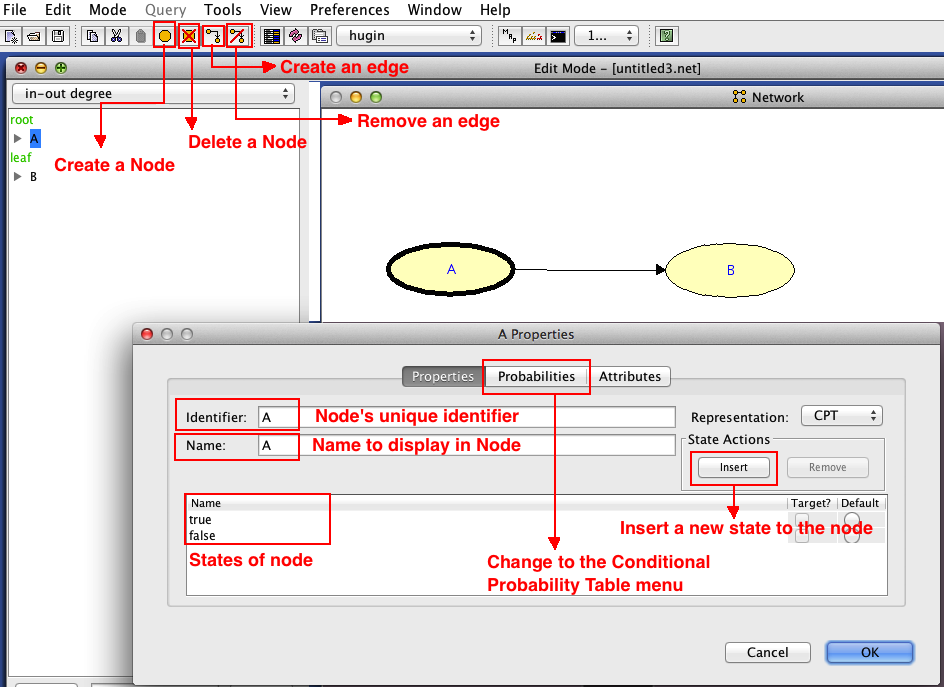}
\caption{SamIam's edit mode default interface.}
\label{fig:edit_mode}
\end{figure}

SamIam provides an intuitive interface for constructing Bayesian Network. There are two modes in SamIam: the query mode (for learning and inferences) and the edit mode (for network structure and definition of conditional probabilities). When SamIam is started, the edit mode appears by default. Figure~\ref{fig:edit_mode} describes the general edit mode interface.

The interface enables the creation/removal of nodes and the creation/removal of edges between nodes. For each node created, there will be a configuration window that can be accessed when the node is \emph{double-clicked}. In this window, one must specify a unique identifier for the node and a name to be displayed in the SamIam interface. Additionally, one also needs to specify which states the node can have. For the scope of this work, we will only have binary random variables, so each node will have exactly two states: one representing the occurrence of the random variable and another representing its absence.

\begin{figure}[h!]
\centering
\includegraphics[scale=0.5]{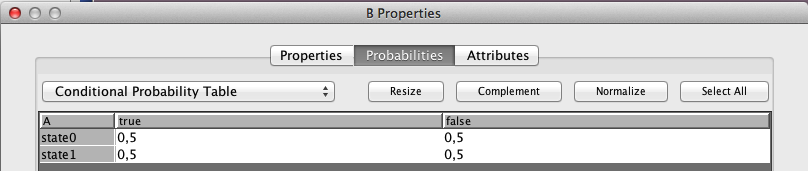}
\caption{SamIam's interface to assign conditional probabilities to random variables.}
\label{fig:cpt}
\end{figure}

The conditional probability table can be accessed by clicking the tab \emph{Probailities}. A windows, similar to the one presented in Figure~\ref{fig:cpt}, will appear.

In this window, a user can manually specify the conditional probabilities of the random variable. by default, Samiam fills these tables using a normal probability distribution, that is, each instance of each node has the same probability of occurring (Pr = 0.5 ).

The buttons \emph{Complement} can be used to automatically assign the last probability value of the table. This takes into account the constraint that the probabilities of an event must sum to one. This way, the user can only manually specify $n - 1$ entries of the table. SamIam computes the remaining probability by subtracting that value with 1: $Pr( N = |n| ) = 1 - \sum_{n = 1}^{ | n | - 1} Pr( N = n )$.

The button \emph{Normalize} normalizes all the entries of the conditional probability table.

\subsection{Learning}

Given a log of events and a graphical structure, SamIam is able to find a statistical model that can automatically estimate the conditional probability tables of the given Bayesian Network. This learning process can be computed using the Maximum Likelihood Estimation (Section~\ref{sec:mle}) if the log of events is complete or using the EM Clustering algorithm if the log of events is incomplete~\citep{Bishop07}.

In the scope of this work, since we were given a complete event log, the process of filling the conditional probability tables was given by the maximum likelihood estimation, that is, by counting the number of times each instance of the log of events was present and then by normalizing to obtain a probability value.

\begin{figure}[h!]
	\parbox{.5\linewidth}{
	\centering
	\includegraphics[scale=0.5]{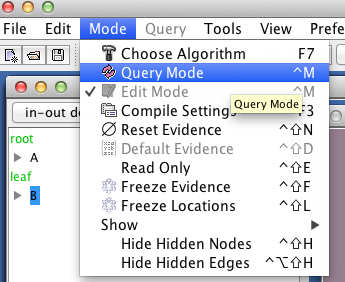}
	\caption{Entering in query mode.}
	\label{fig:querymode1}
	}
	\hfill
	\parbox{.5\linewidth}{
	\centering
	\includegraphics[scale=0.5]{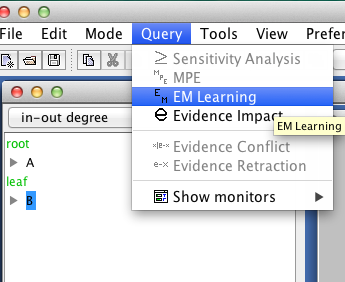}
	\caption{Entering in the learning menu.}
	\label{fig:querymode2}	
	}
\end{figure}

SamIam can automatically do this in the query mode. In the main SamIam interface, one can select the \emph{query mode} just like presented in Figure~\ref{fig:querymode1}. To go into the learning menu, one needs to find the option \emph{EM Learning} (Figure~\ref{fig:querymode2}).

\begin{figure}[h!]
\centering
\includegraphics[scale=0.6]{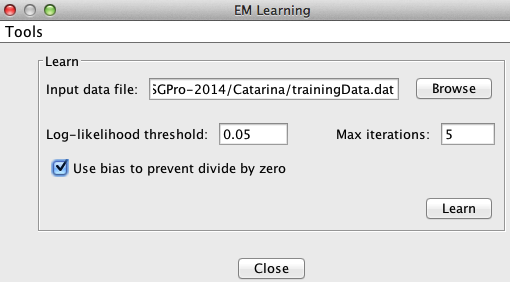}
\caption{SamIam learning menu. }
\label{fig:samiam_learning}
\end{figure}

Under the \emph{EM Learning} menu, the user is presented with a window that asks for a training file, a probability threshold, the maximum number of iterations that the algorithm should perform and if the learning algorithm should ignore entries that lead to divisions by zero. Figure~\ref{fig:samiam_learning} illustrates these options.

In Figure~\ref{fig:samiam_learning}, the field \emph{Max iterations} corresponds to the total number of iterations that the \emph{EM Clustering} should perform in case the algorithm does not converge. For the scope of this work, this entry is irrelevant since we are dealing with fully observed log of events. Consequently, the EM Clustering will collapse to the Maximum Likelihood Estimate.

The field \emph{Log-likelihood threshold} is also used in the scope of the EM Clustering. This threshold specifies that the algorithm will converge when the change in the log-likelihood function falls bellow a certain threshold. It is a common practice in the literature to set this value to $0.05$~\citep{Bishop07,Koller09}.

The option \emph{Use bias to prevent divisions by zero} should always be used, otherwise the Maximum Likelihood Estimate formula will try to perform a division by zero when it tries to compute the probability of an instance that does not exist in the training set.

\begin{figure}[h!]
	\centering
	\includegraphics[scale=0.33]{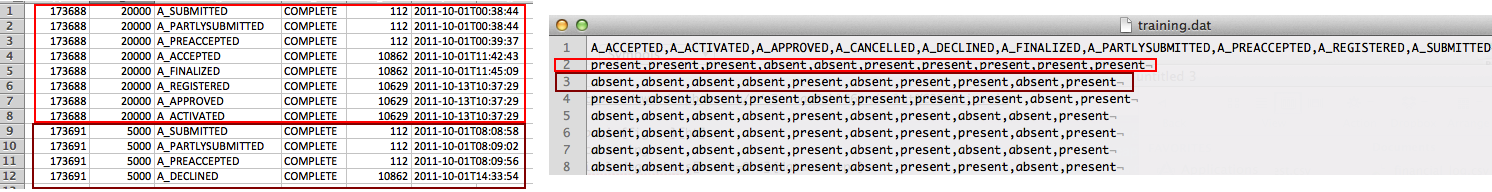}
	\caption{Entering in the learning menu.}
	\label{fig:training}	
\end{figure}

In process mining, a training set consists in a portion of the log of events that is used to fit (train) a model for prediction of values. In the scope of this work, a training set will consist of 70\% randomized entries of the log of events. The format of the training file contains the names of all random variables (nodes) in the first line. The remaining lines of the file correspond to the instances of the nodes that are specified in the log of events. In this work, we modeled binary random variables with the instances \emph{present} to represent the occurrence of a task in the business process and \emph{absent} to represent the non-existence of the task. Figure~\ref{fig:training} shows the log of events (left) and the conversion of one instance of the log of events into a training file with the SamIan format (right).

\begin{figure}[h!]
	\parbox{.5\linewidth}{
	\centering
	\includegraphics[scale=0.6]{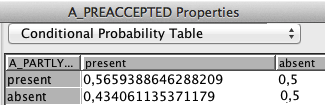}
	\caption{Learned conditinal probability table.}
	\label{fig:cpt_learned}
	}
	\hfill
	\parbox{.5\linewidth}{
	\centering
	\includegraphics[scale=0.6]{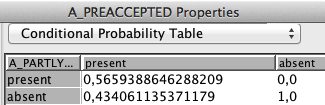}
	\caption{Corrected conditional probability table denoting mutual exclusion between the nodes.}
	\label{fig:cpt_corrected}	
	}
\end{figure}

After SamIam learns the conditional probability tables, it is necessary to correct some semantics of the network. More specific the inclusion of mutually exclusive relationships. For instance, Figure~\ref{fig:cpt_learned} presents a conditional probability table that was automatically learned by SamIam. As one can see, when the node A\_PARTLYSUBMITTED is absent, SamIam did not update the normal probability distribution, so the probabilities $0.5$ remained in the conditional probability table. This means that there were no events in the log that did not have an instance of the A\_PARTLYSUBMITTED node. This happens, because in process mining, the activities that are performed are usually mutually exclusive, unless a special structure is used to say the contrary. In order to correct these probabilities, such that the mutual exclusion is captured, one just needs to fill the conditional probability table just like illustrated in Figure~\ref{fig:cpt_corrected}. When the preceding node is \emph{absent}, then the posterior nodes should also become \emph{absent}.

\section{Case Study: Loan Application}\label{sec:case_study}

The event log that we use in this work is taken from a Dutch Financial Institute\footnote{\url{http://www.win.tue.nl/bpi/2012/challenge
}}. The event log represents a loan application belonging to a global financial organization, in which a customer requests a certain amount of money. The process is composed of three different sub processes. The first letter of each task corresponds to an identifier of the sub process it belongs to. The tasks that start with letter $A\_$ correspond to states of the application. The tasks that start with letter $O\_$ correspond to offers belonging to the application. And the tasks that start with letter $W\_$ correspond to the work item belonging to the application.

The general scenario is as follows. There is a webpage that enables the submission of loan applications. A customer selects a certain amount of money and then submits his request. Then, the application performs some automatic tasks and checks if an application is eligible. If it is eligible, then the customer is sent an offer by mail.  After this offer is received, it will be evaluated. In case of any missing information, the offer goes back to the client and is again evaluated until all the required information is gathered. A final evaluation is done to the application. Finally, the application is approved and activated.

The log contains $262 200$ events and $13 087$ cases. The statistics of the log of events is summarised in Table~\ref{tab:data}.

\begin{table}[h!]
\centering
\begin{tabular}{ l | c | l | c }
{\bf Event}					& {\bf Num Occurrences}  & {\bf Event}	 & {\bf Num Occurrences}\\
\hline
A\_SUBMITTED				& 13 087 				& W\_Nabellen incomplete dossiers	& 11 407	\\
A\_PARTLYSUBMITTED			& 13 087				& W\_Valideren aanvraag			& 7 895\\
A\_PREACCEPT				& 7 367				& W\_Afhandelen leads				& 5898\\
A\_CANCELLED				& 2 807				& W\_Beoordelen fraude				& 270	\\
A\_APPROVED				& 2 246				&  W\_Wijzigen contractgegevens		& 0 		\\
A\_REGISTERED				& 2 246				& O\_ACCEPTED					& 2 243 	\\
A\_ACTIVATED				& 2 246				& O\_SELECTED					& 7 030 	 \\
A\_DECLINED					& 7 635				& O\_CREATED					& 7 030	 \\
A\_FINALIZED					& 5 015 				& O\_SENT						& 7 030\\
A\_ACCEPTED				& 5 113				& O\_SENT\_BACK					& 3 454 	 \\
W\_Completeren aanvraag		& 23 967				& O\_CANCELLED					& 3 655	\\
W\_Nabellen offertes			& 22 976				& O\_DECLINED					& 802	\\
\end{tabular}
\caption{Summary of the statistics of the Loan Application event log. Only COMPLETE events were taken into account. }
\label{tab:data}
\end{table}
\subsection{Converting the Log of Events into a SamIam Bayesian Nework}

In this work, a Java program was made that received as input the log of events in \emph{csv} format and returned a Bayesian Network in a special file format that can be readable by the SamIam toolkit.  The program parsed every line of the log of events and grouped all activities that were complete and that belonged to the same instance (had the same caseId). The program automatically created a graph in a matrix form representation and computed the frequency of the connections between nodes.

\begin{figure}[h!]
	\parbox{.6\linewidth}{
	\centering
	\includegraphics[scale=0.4]{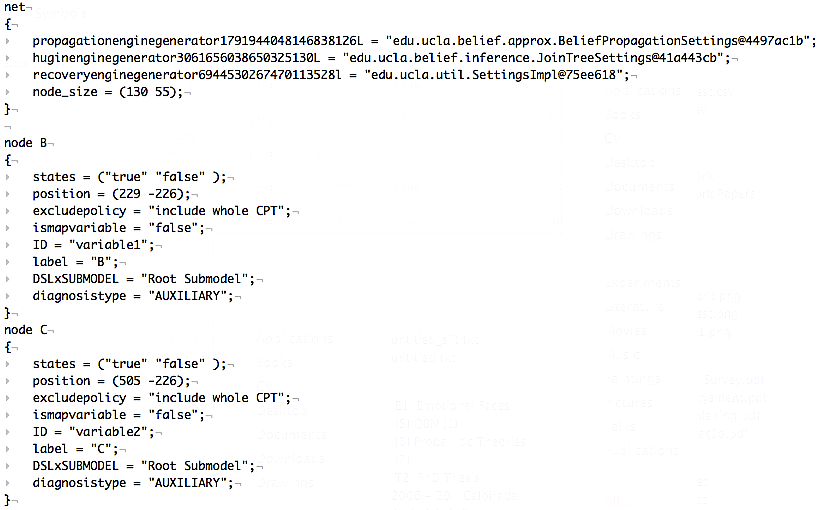}
	\caption{Example of a SamIam network file.}
	\label{fig:samiam1}
	}
	\hfill
	\parbox{.3\linewidth}{
	\centering
	\includegraphics[scale=0.48]{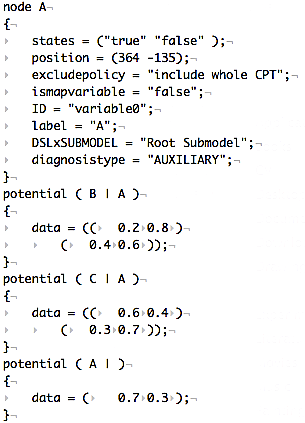}
	\caption{Example of a SamIam network file.}
	\label{fig:samiam2}	
	}
\end{figure}

Given this matrix form graph representation, another Java program was made in order to convert this matrix into a network file recognized by SamIam. Figures~\ref{fig:samiam1} and~\ref{fig:samiam2} show an example of a network file readable by SamIam. This example shows a network of the following form: $C \leftarrow A \rightarrow B$.

\begin{figure}[h!]
\includegraphics[scale=0.5]{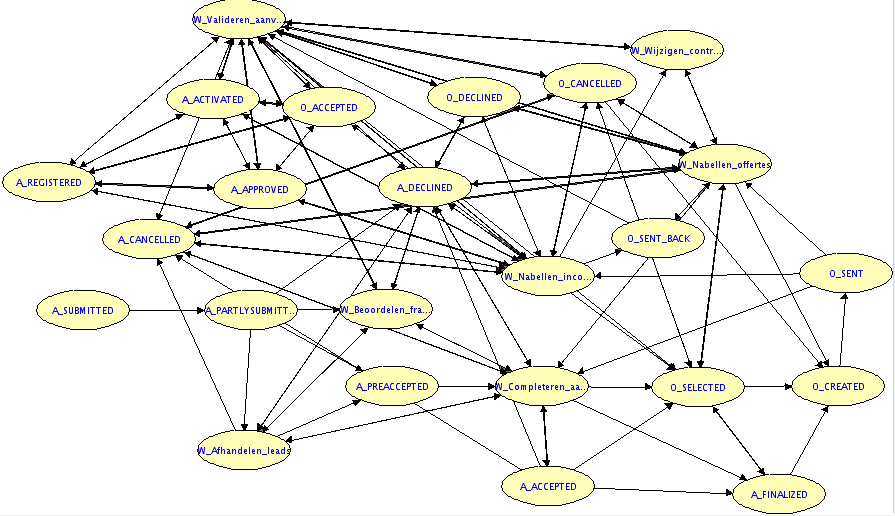}
\caption{Full representation of the Loan Application Bayesian Network.}
\label{fig:full_log}
\end{figure}

In a first attempt, we mapped the entire log of events into a Bayesian Network. However, the full log contained many tasks (about 24 random variables) and turned the process too big and complex to analyse. Figure~\ref{fig:full_log} shows the network directly extracted from the log of events. The cycles that are present in this network were already expected, since the log of events contain many events that require cycles. Later in this work, we will specify an heuristic to remove such cyclic structures and turn any network into an acyclic directed graph. 

Since the network in Figure~\ref{fig:full_log} was too complex, we decided to choose only the nodes concerned with the $A\_$ tasks of the log of events, just like it was done in the works of~\citep{Adriansyah12,Bautista12,Bose12,Kang12}.

The resulting Bayesian Network was smaller, containing only $10$ random variables. We then altered the Bayesian Network in order to add mutually exclusive relationships between the nodes A\_DECLINED and A\_CANCELLED and between the nodes A\_APPROVED, A\_DECLINED and A\_CANCELLED.

\begin{figure}[h!]
\includegraphics[scale=0.55]{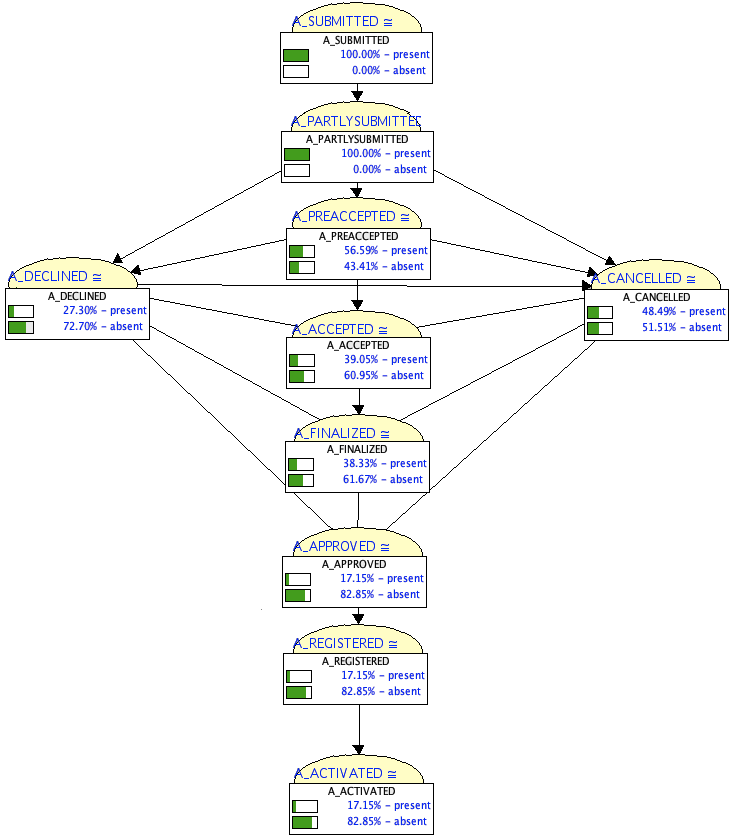}
\caption{Bayesian Network representation of the loan application. Only A\_ nodes were taken into account. Manually added mutually exclusive relationships between nodes A\_DECLINED and A\_CANCELLED and between nodes A\_APPROVED, A\_DECLINED and A\_CANCELLED.}
\label{fig:loan_net}
\end{figure}

\begin{figure}[h!]
\centering
	\includegraphics[scale=0.35]{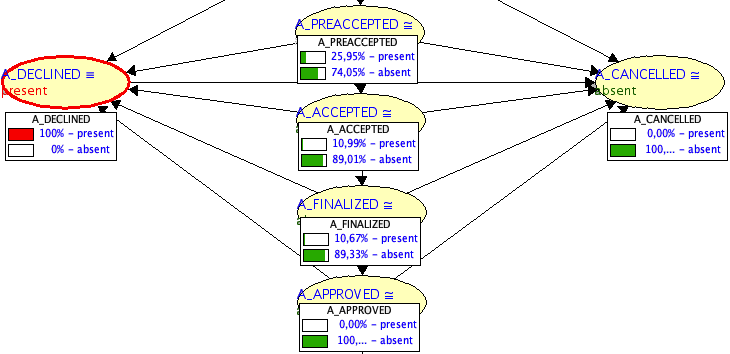}
	\caption{Mutual exclusion between nodes A\_DECLINED, A\_CANCELLED and A\_APPROVED.}
	\label{fig:loan_mutual1}
\end{figure}

\begin{figure}[h!]
	\parbox{.42\linewidth}{
	\centering
	\includegraphics[scale=0.35]{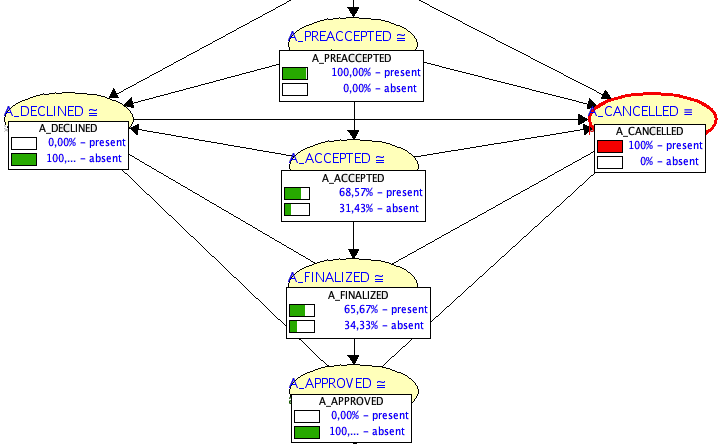}
	\caption{Mutual exclusion between nodes A\_DECLINED, A\_CANCELLED and A\_APPROVED.}
	\label{fig:loan_mutual2}	
	}
	\hfill
	\parbox{.42\linewidth}{
	\centering
	\includegraphics[scale=0.35]{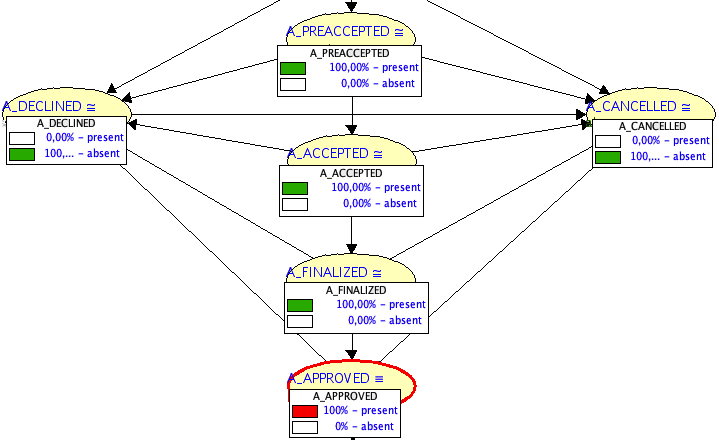}
	\caption{Mutual exclusion between nodes A\_DECLINED, A\_CANCELLED and A\_APPROVED.}
	\label{fig:loan_mutual3}	
	}
\end{figure}

The mutually exclusive relation between the nodes A\_DECLINED and A\_CANCELLED is straightforward. A loan application cannot be both declined and cancelled. Additionally, if an application is known to be declined, then the probability of being cancelled will be zero and vice-versa. Figure~\ref{fig:loan_net} presents this network and Figures~\ref{fig:loan_mutual1} to~\ref{fig:loan_mutual3} illustrate the mutual exclusion between nodes.

In order to compare our model with other works in the literature, we also created a Markov Chain from the same log of events (Section~\ref{sec:mc}). We then computed the probability of each sequence of the test set occurring in the Bayesian Network and in the Markov Chain and then compared the results. Section~\ref{sec:results} presents the main outcomes of these experiments.

\subsection{Converting the Log of Events into a Markov Chain}

\begin{figure}[h!]
\includegraphics[scale=0.55]{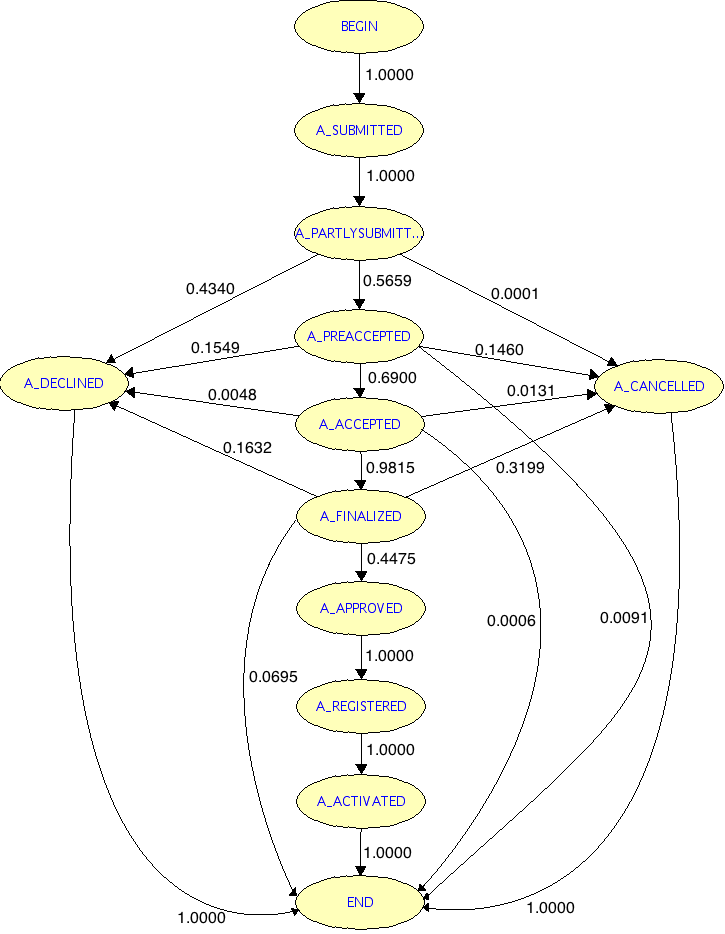}
\caption{Markov Chain representation of the loan application.}
\label{fig:mc}
\end{figure}

As already mentioned, we also developed a Markov Chain by a script in Python with the same training set used to generate the Bayesian Network. The transition probabilities of the Markov Chain were computed by simply counting the number of occurrences of each sequence of events and then by normalizing to obtain a probability value. Figure~\ref{fig:mc} shows the computed Markov Network.

\subsection{Results} \label{sec:results}

After defining the structure of the Bayesian Network for the loan application, we generated a training set in which we randomly selected 70\% of cases in the event log as training set and then, we used the remaining $30\%$ as a test set to validate our model.

The training set was given as input to SamIam in order to learn the conditional probability tables. Then, to test the application, a MatLab program was developed in order to perform probabilistic inferences. Basically, the MatLab program received as input the SamIam's network file and returned a Bayesian Network structure. From this program, we were able to compute full joint probability distributions and marginal probabilities. Another Java program received as input the test set and was able to validate the model. The validation was performed as follows: we computed the probability of some events occurring in the test set and then we compared this value with the probability given in the trained Bayesian Network. Tables~\ref{tab:res1} to~\ref{tab:res7} show the results obtained for different queries both in the test set and the training set.

\begin{table}[h!]
\centering
\begin{tabular}{l | c | c  | c }
{\bf Probability}						& {\bf Test Set}	& {\bf Training Set}		& {\bf ERROR \%} \\
\hline
Pr( A\_CANCELLED = present )			& 0.0000			& 0.0000 				& 0.0000\\
Pr( A\_ACTIVATED  = present )			& 0.0000			& 0.0000 				& 0.0000\\
Pr( A\_ACCEPTED = present )			& 0.1063 			& 0.1099 				& 0.3592\\
Pr( A\_FINALIZED = present )				& 0.1010 			& 0.1067  			& 0.5685\\
Pr( A\_PREACCEPT = present )			& 0.2307 			& 0.2595 				& 2.8799\\
Pr( A\_SUBMITTED = present )			& 1.0000 			& 1.0000 				& 0.0000\\
\hline
\end{tabular}
\caption{Results obtain when the node A\_DECLINED  = present was observed.}
\label{tab:res1}
\end{table}

\begin{table}[h!]
\centering
\begin{tabular}{l | c | c | c | c }
{\bf Probability}						& {\bf Test Set}	& {\bf Training Set}		& {\bf ERROR \%} \\
\hline
Pr( A\_DECLINED  = present )				& 0.0000 		& 0.0000 		& 0.0000 \\
Pr( A\_ACTIVATED  = present )			& 0.0000 		& 0.0000 		& 0.0000 \\
Pr( A\_ACCEPTED = present )			& 0.6098 		& 0.6857 		& 7.5916 \\
Pr( A\_FINALIZED = present )				& 0.5882 		& 0.6567 		& 6.8532 \\
Pr( A\_PREACCEPT = present )			& 1.0000 		& 1.0000 		& 0.0000\\
Pr( A\_SUBMITTED = present )			& 1.0000 		& 1.0000 		& 0.0000\\
\hline
\end{tabular}
\caption{Results obtain when the node A\_CANCELLED  = present was observed.}
\label{tab:res2}
\end{table}

\begin{table}[h!]
\centering
\begin{tabular}{l | c | c | c | c }
{\bf Probability}						& {\bf Test Set}	& {\bf Training Set}		& {\bf ERROR \%} \\
\hline
Pr( A\_DECLINED  = present )				& 0.5773 		& 0.5860 		& 0.8715\\
Pr( A\_ACTIVATED  = present )			& 0.1719 		& 0.1715 		& 0.0387\\
Pr( A\_ACCEPTED = present )			& 0.3911 		& 0.3905 		& 0.0638\\
Pr( A\_FINALIZED = present )				& 0.3830 		& 0.3833 		& 0.0310\\
Pr( A\_PREACCEPT = present )			& 0.5559 		& 0.5659 		& 1.0005\\
Pr( A\_CANCELLED = present )			& 0.2238 		& 0.1315 		& 9.2335\\
\hline
\end{tabular}
\caption{Results obtain when the node A\_SUBMITTED  = present was observed.}
\label{tab:res3}
\end{table}

\begin{table}[h!]
\centering
\begin{tabular}{l | c | c | c | c }
{\bf Probability}						& {\bf Test Set}	& {\bf Training Set}		& {\bf ERROR \%} \\
\hline
Pr( A\_DECLINED  = present )				& 0.2396 		& 0.2687 		& 2.9121 \\
Pr( A\_ACTIVATED  = present )			& 0.3092 		& 0.3030 		& 0.6208 \\
Pr( A\_ACCEPTED = present )			& 0.7036		& 0.6900 		& 1.3619 \\
Pr( A\_FINALIZED = present )				& 0.6890 		& 0.6773 		& 1.1660 \\
Pr( A\_SUBMITTED = present )			& 1.0000 		& 1.0000 		& 0.0000 \\
Pr( A\_CANCELLED = present )		 	& 0.4027 		& 0.2324 		& 17.0257\\
\hline
\end{tabular}
\caption{Results obtain when the node A\_PREACCEPT  = present was observed.}
\label{tab:res4}
\end{table}

\begin{table}[h!]
\centering
\begin{tabular}{l | c | c | c | c }
{\bf Probability}						& {\bf Test Set}	& {\bf Training Set}		& {\bf ERROR \%} \\
\hline
Pr( A\_DECLINED  = present )				& 0.1523 		& 0.1632 		& 1.0939 \\
Pr( A\_ACTIVATED  = present )			& 0.4488 		& 0.4475 		& 0.1303\\
Pr( A\_ACCEPTED = present )			& 1.0000 		& 1.0000 		& 0.0000\\
Pr( A\_PREACCEPT = present )		 	& 1.0000 		& 1.0000 		& 0.0000\\
Pr( A\_SUBMITTED = present )			& 1.0000 		& 1.0000 		& 0.0000\\
Pr( A\_CANCELLED = present )			& 0.3438 		& 0.2254 		& 11.8350\\
\hline
\end{tabular}
\caption{Results obtain when the node A\_FINALIZED  = present was observed.}
\label{tab:res5}
\end{table}

\begin{table}[h!]
\centering
\begin{tabular}{l | c | c | c | c}
{\bf Probability}						& {\bf Test Set}	& {\bf Training Set}		& {\bf ERROR \%} \\
\hline 
Pr( A\_DECLINED  = present ) 			& 0.1569 		& 0.1649 		& 0.7999\\
Pr( A\_ACTIVATED  = present ) 			& 0.4395 		& 0.4392 		& 0.0253\\
Pr( A\_FINALIZED = present ) 			& 0.9792		& 0.9815		& 0.2333\\
Pr( A\_PREACCEPT = present ) 		& 1.0000 		& 1.0000 		& 0.0000\\
Pr( A\_SUBMITTED = present )			& 1.0000 		& 1.0000 		& 0.0000\\
Pr( A\_CANCELLED = present )			& 0.3490 		& 0.2310 		& 11.7958\\
\hline
\end{tabular}
\caption{Results obtain when the node A\_ACCEPTED  = present was observed.}
\label{tab:res6}
\end{table}

\begin{table}[h!]
\centering
\begin{tabular}{l | c | c | c | c }
{\bf Probability}						& {\bf Test Set}	& {\bf Training Set}		& {\bf ERROR \%} \\
\hline
Pr( A\_DECLINED  = present ) = 			& 0.0000 		& 0.0000 		& 0.0000\\
Pr( A\_ACCEPTED = present ) = 			& 1.0000 		& 1.0000 		& 0.0000\\
Pr( A\_FINALIZED = present ) = 			& 1.0000 		& 1.0000 		& 0.0000\\
Pr( A\_PREACCEPT = present ) = 		& 1.0000 		& 1.0000 		& 0.0000\\
Pr( A\_SUBMITTED = present ) = 			& 1.0000 		& 1.0000 		& 0.0000\\
Pr( A\_CANCELLED = present ) = 			& 0.0000 		& 0.0000 		& 0.0000\\
\hline
\end{tabular}
\caption{Results obtain when the node A\_ACTIVATED  = present was observed.}
\label{tab:res7}
\end{table}

The overall results show that the Bayesian Network learned from the log of events is a good approach for process mining, since the errors obtained were very low. The most significant errors come associated with the node A\_CANCELLED. For instance, in Table~\ref{tab:res4}, the probability Pr( A\_CANCELLED = present $|$ A\_PREACCEPT ) achieved an error of $17 \%$. One possible explanation can be given by the mutual exclusivities that were given to this node. Since in Bayesian Networks, all nodes depend of each other, then by adding new relationships to the nodes, we are introducing some non-trivial effects in the model.

Another experiment made was to compare the proposed Bayesian Network with a Markov Chain. We trained a Markov Chain in the same way we did for the Bayesian Network. 

\begin{table}
\centering
\begin{tabular}{ l l | l l }
{\bf Processes}			& {\bf Process Encoding} 	& {\bf Processes}		& {\bf Process Encoding}\\
\hline
A\_SUBMITTED			&  A\_SUB				& A\_APPROVED		&  A\_APPR \\
A\_PARTLYSUBMITTED		&  A\_PART				& A\_REGISTERED		&  A\_REG \\
A\_PREACCEPT			&  A\_PRE				& A\_ACTIVATED		&  A\_ACT\\
A\_ACCEPTED			&  A\_ACC				& A\_DECLINED		&  A\_DEC \\
A\_FINALIZED				&  A\_FIN					& A\_CANCELLED		&  A\_CAN\\
\hline
\end{tabular}
\caption{Encodings of the nodes used in the Bayesian Network.}
\label{tab:encoding}
\end{table}

In order to validate both approaches, we leveraged on the test set and computed the probability of each sequence occurring in a Bayesian Network and in a Markov Chain. In the end, those probabilities were weighted with the number of occurrences of each sequence in the test set. The results obtained are discriminated in Table~\ref{tab:resfinal}.

\begin{table}[h!]
\resizebox{\columnwidth}{!} {
\begin{tabular}{l | c | c | c  | c }
{\bf Chain}			& {\bf  Occ. Test Set} & {\bf BN} 	 & {\bf MC}  & {\bf ERROR \% } \\
\hline
{\bf A\_SUB $\rightarrow$ A\_PART $\rightarrow$ A\_PRE }				 			& 22 	& 0.046426  	& 0.0051 		& 4.13 \%	\\
{\bf A\_SUB $\rightarrow$ A\_PART $\rightarrow$ A\_PRE  $\rightarrow$ A\_ACC } 		& 1		& 0.00154		& 0.0002  	& 0.13 \% \\
{\bf A\_SUB $\rightarrow$ A\_PART $\rightarrow$ A\_PRE  $\rightarrow$ A\_ACC  $\rightarrow$ A\_FIN } & 83 	& 0.0627 	& 0.0266 	& 3.61 \% \\
{\bf A\_SUB $\rightarrow$ A\_PART $\rightarrow$ A\_DEC}						& 1744 	& 0.433843 	& 0.4340 	& 0.01\% \\
{\bf A\_SUB $\rightarrow$ A\_PART $\rightarrow$ A\_PRE  $\rightarrow$ A\_DEC}	& 282	& 0.046369	& 0.0877	& 4.13 \% \\
{\bf A\_SUB $\rightarrow$ A\_PART $\rightarrow$ A\_PRE  $\rightarrow$ A\_ACC $\rightarrow$ A\_DEC} & 12	& 0.000534	& 0.0019 & 0.13 \% \\
{\bf A\_SUB $\rightarrow$ A\_PART $\rightarrow$ A\_PRE   $\rightarrow$  A\_ACC   $\rightarrow$  A\_FIN  $\rightarrow$   A\_DEC}	& 229 & 0.02363 & 0.0626	& 3.90 \\
{\bf A\_SUB  $\rightarrow$ A\_PART  $\rightarrow$ A\_PRE  $\rightarrow$ A\_CAN}	& 343	& 0.041347	& 0.0826	& 4.13 \%	\\
{\bf A\_SUB  $\rightarrow$ A\_PART  $\rightarrow$  A\_PRE  $\rightarrow$ A\_ACC  $\rightarrow$ A\_CAN}	& 19	& 0.003809	& 0.0051	& 0.13 \% \\
{\bf A\_SUB $\rightarrow$  A\_PART $\rightarrow$ A\_PRE $\rightarrow$  A\_ACC $\rightarrow$  A\_FIN $\rightarrow$ A\_CAN}	& 517	& 0.0864	& 0.1226	& 3.62 \% \\
{\bf A\_SUB  $\rightarrow$  A\_PART  $\rightarrow$ A\_PRE  $\rightarrow$  A\_ACC  $\rightarrow$  A\_FIN  $\rightarrow$ A\_APPR  $\rightarrow$  A\_REG  $\rightarrow$ A\_ACT}		& 675 	& 0.1715		& 0.1715		& 0.0000 \% \\	
\hline
{\bf Total}		& 		& {\bf 0.2435}		& {\bf 0.2561} 			& {\bf 1.2674 \%	}\\
\end{tabular}
}
\caption{Comparison of a Bayesian Network (BN) and a Markov Chain (MC) for process mining. The Error \% was computed in the following way: $\left|  BN - MC  \right|*100$}.
\label{tab:resfinal}
\end{table}

Table~\ref{tab:resfinal} shows that the probabilities computed in  a Bayesian Network are almost identical to the ones computed by the Markov Chain. Individually, the probabilities of computing the sequences in the test set did not have an error percentage superior to $4.13 \%$, which is statistically insignificant given the total amount of data tested. Moreover, the overall error percentage between the proposed Bayesian Network and the Markov Chain was around $1.2674 \%$, which is also statistically insignificant. This means that the Bayesian Networks have a similar performance as a Markov Chain. Consequently, one can conclude that Bayesian Networks are also good approaches to model business processes, with the advantage of being able to represent uncertainty (computing probabilities of tasks that we do not know if occurred).

\subsection{Queries }

As already mentioned, one of the capabilities of Bayesian Networks for process mining is their ability to deal with uncertainty. They enable the analysis of tasks that are not known to occur. For instance, for the Loan Application Bayesian Network, one can be interested in analyzing the probability of the business process ending successfully by only knowing that a couple of tasks were observed to occur. Combining this ability with SamIam's graphical capabilities will enable a fast analysis of business processes as well as risk management. 

\begin{figure}[h!]
\centering
\includegraphics[scale=0.5]{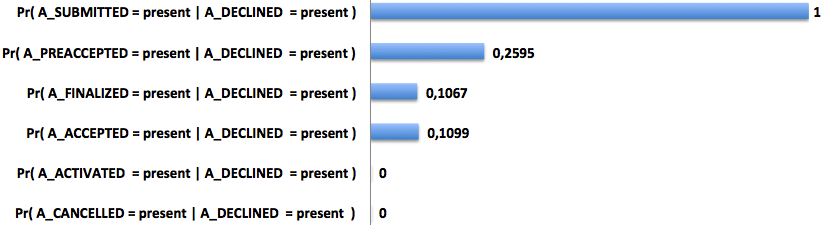}
\caption{Analysis of the probabilities of reaching some nodes of the Loan Application Bayesian Network, when it is known that the application was \emph{declined}.}
\label{fig:declined}
\end{figure}

Figure~\ref{fig:declined} shows the probabilities of some nodes of the Loan Application Bayesian Network, when it is only known that the application was declined, that is, the node A\_DECLINED was observed to occur. From this analysis, one can conclude that the majority of the applications that are declined have a high probability of reaching the state A\_PREACCEPT. Moreover, if an application is declined, then the nodes A\_ACTIVATED and A\_CANCELLED are never reached.

\begin{figure}[h!]
\centering
\includegraphics[scale=0.55]{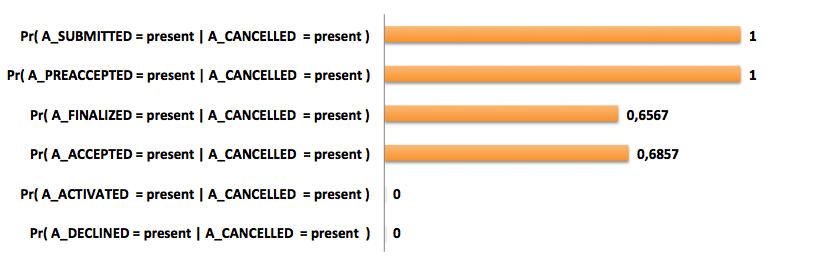}
\caption{Analysis of the probabilities of reaching some nodes of the Loan Application Bayesian Network, when it is known that the application was \emph{cancelled}.}
\label{fig:cancelled}
\end{figure}

Another example is given by Figure~\ref{fig:cancelled}. When it is known that the application ended up in a cancelled state, then one can estimate with a $100\%$ probability that the process reached the task A\_PREACCEPT and never reached the tasks A\_DECLINED and A\_ACTIVATED. Moreover, there is a high probability that the application was cancelled during the tasks A\_ACCEPTED and A\_FINALIZED.

\begin{figure}[h!]
\centering
\includegraphics[scale=0.55]{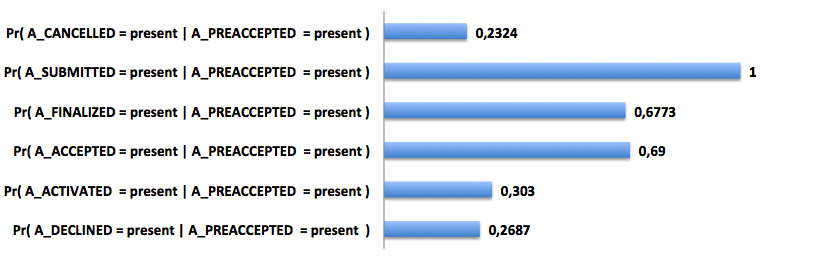}
\caption{Analysis of the probabilities of reaching some nodes of the Loan Application Bayesian Network, when it is known that the application was \emph{pre accepted}.}
\label{fig:preaccepted}
\end{figure}

The maximum uncertainty in the loan application business process is given when one only knows that the process was started, which happens when the task A\_SUBMITTED is observed to occur. In this situation, the proposed Bayesian Network estimates that there is a high probability of the process going to the task A\_PREACCEPT or being declined (A\_DECLINED). If one chooses task A\_PREACCEPT, then from Figure~\ref{fig:preaccepted} one can conclude that there is a high probability that the process will be either accepted or finalised.

\begin{figure}[h!]
\centering
\includegraphics[scale=0.55]{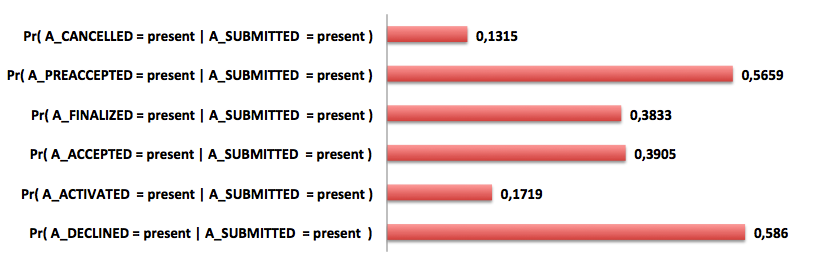}
\caption{Analysis of the probabilities of reaching some nodes of the Loan Application Bayesian Network, when it is known that the application was \emph{submitted}.}
\label{fig:submitted}
\end{figure}

\section{Conclusion and Future Work}\label{sec:conclusions}

In this work, we propose the usage of Bayesian Networks as a new approach to represent business processes automatically extracted from event logs.

In a first step, we extracted the relationships between nodes from the log of events and then used this log to train and validate the proposed Bayesian Network.

Experiments made over a Loan Application Case study suggest that Bayesian Networks have the same performance as Markov Chains, so they are good models to make accurate predictions about sequences of events in the scope of process mining.

Moreover, by modelling a business process through Bayesian Networks, one is able to take advantage of the ability of these structures to deal with uncertainty. More specifically, Bayesian Networks enable the reconstruction of a flow by only taking into account partial observations in the business process. 

As for future work, it would be interesting to extend the capabilities of Bayesian Networks to learn from incomplete logs of events. One could train such network using the EM Clustering in order to find an approximate probability distribution for the occurrence of the tasks. Moreover, together with SamIam, one could try to estimate the most probable sequences of business processes using the probabilities learned from the incomplete log.
\bibliographystyle{agsm}

\end{document}